\documentclass[letterpaper, 10 pt, conference]{ieeeconf}  % Comment this line out if you need a4paper

\IEEEoverridecommandlockouts                              % This command is only needed if 
\usepackage{graphicx}

\title{\LARGE \bf
3D-MOV: Audio-Visual LSTM Autoencoder for 3D Reconstruction of Multiple Objects from Video
}

\author{Justin Wilson$^{1}$ and Ming C. Lin$^{2}$% <-this % stops a space
\thanks{$^{1}$Justin Wilson, Department of Computer Science,
        University of North Carolina at Chapel Hill, 201 S. Columbia St., Chapel Hill, NC 27599, USA
        {\tt\small wilson@cs.unc.edu}}%
\thanks{$^{2}$Ming C. Lin is Dr. Barry Mersky and Capital One Endowed Professor and former Elizabeth Stevinson Iribe Chair of Computer Science, University of Maryland at College Park, 8125 Paint Branch Drive, College Park, MD 20742, USA
        {\tt\small lin@cs.umd.edu and lin@cs.unc.edu}}%
}

\begin{document}

\maketitle
\thispagestyle{empty}
\pagestyle{empty}

%%%%%%%%%%%%%%%%%%%%%%%%%%%%%%%%%%%%%%%%%%%%%%%%%%%%%%%%%%%%%%%%%%%%%%%%%%%%%%%%
\begin{abstract}

3D object reconstructions of transparent and concave structured objects, with inferred material properties, remains an open research problem for robot navigation in unstructured environments. In this paper, we propose a multimodal single- and multi-frame neural network for 3D reconstructions using audio-visual inputs. Our trained reconstruction LSTM autoencoder 3D-MOV accepts multiple inputs to account for a variety of surface types and views. Our neural network produces high-quality 3D reconstructions using voxel representation. Based on Intersection-over-Union (IoU), we evaluate against other baseline methods using synthetic audio-visual datasets ShapeNet and Sound20K with impact sounds and bounding box annotations. To the best of our knowledge, our single-
and multi-frame model is the first audio-visual reconstruction neural network for 3D geometry and material representation.

\end{abstract}

%%%%%%%%%%%%%%%%%%%%%%%%%%%%%%%%%%%%%%%%%%%%%%%%%%%%%%%%%%%%%%%%%%%%%%%%%%%%%%%%
\section{INTRODUCTION}
\label{Introduction}

Deep neural networks trained on single- or multi-view images have enabled 3D reconstruction of objects and scenes using RGB and RGBD approaches for robotics and other 3D vision-based applications. These models generate 3D geometry volumetrically~\cite{DBLP:journals/corr/BoscainiMRB16,choy20163d,DBLP:journals/corr/abs-1801-09710} and in the form of point clouds~\cite{han2019multiangle,qi2016pointnet,qi2017pointnet}. With these reconstructions, additional networks have been developed to use the 3D geometry as inputs for object detection, classification, and segmentation in 3D environments~\cite{DBLP:journals/corr/abs-1803-10091,DBLP:journals/corr/abs-1711-08488}. However, existing methods still encounter a few challenging scenarios for 3D shape reconstruction~\cite{DBLP:journals/corr/BoscainiMRB16}.

One such challenge is occlusion in cluttered environments with multiple agents/objects in a scene. Another is spatial resolution. Volumetric methods such as voxelized reconstructions~\cite{Maturana-2015-6018} are primarily limited by resolution. Point cloud representations of shape avoid issues of grid resolution, but instead need to cope with issues of point set size and approximations. Existing methods also are challenged by transparent and highly reflective or textured surfaces. Self-occlusions and occlusions from other objects can also hinder image-based networks, necessitating the possible adoption of multimodal neural networks.

To address these limitations, we propose to use audio-visual input for 3D shape and material reconstruction. A single view of an object is insufficient for 3D reconstruction as only one projection of the object can be seen, while multi-view input does not intrinsically model the spatial relationships between views. By providing a temporal sequence of video frames, we strengthen the relationships between views, aiding reconstruction. We also include audio as an input, in particular, \emph{impact sounds} resulting from interactions between the object to be reconstructed and the surrounding environment. Impact sounds provide information about the material and internal structure of an object, offering complementary cues to the object's visual appearance. We choose to represent our final 3D shape using voxel representation due to their state-of-the-art performance in classification tasks.
To the best of our knowledge, our audio-visual network is the first to reconstruct multiple 3D objects from a single video.

\begin{figure*}[thpb]
  \centering
  \includegraphics[width=1\textwidth]{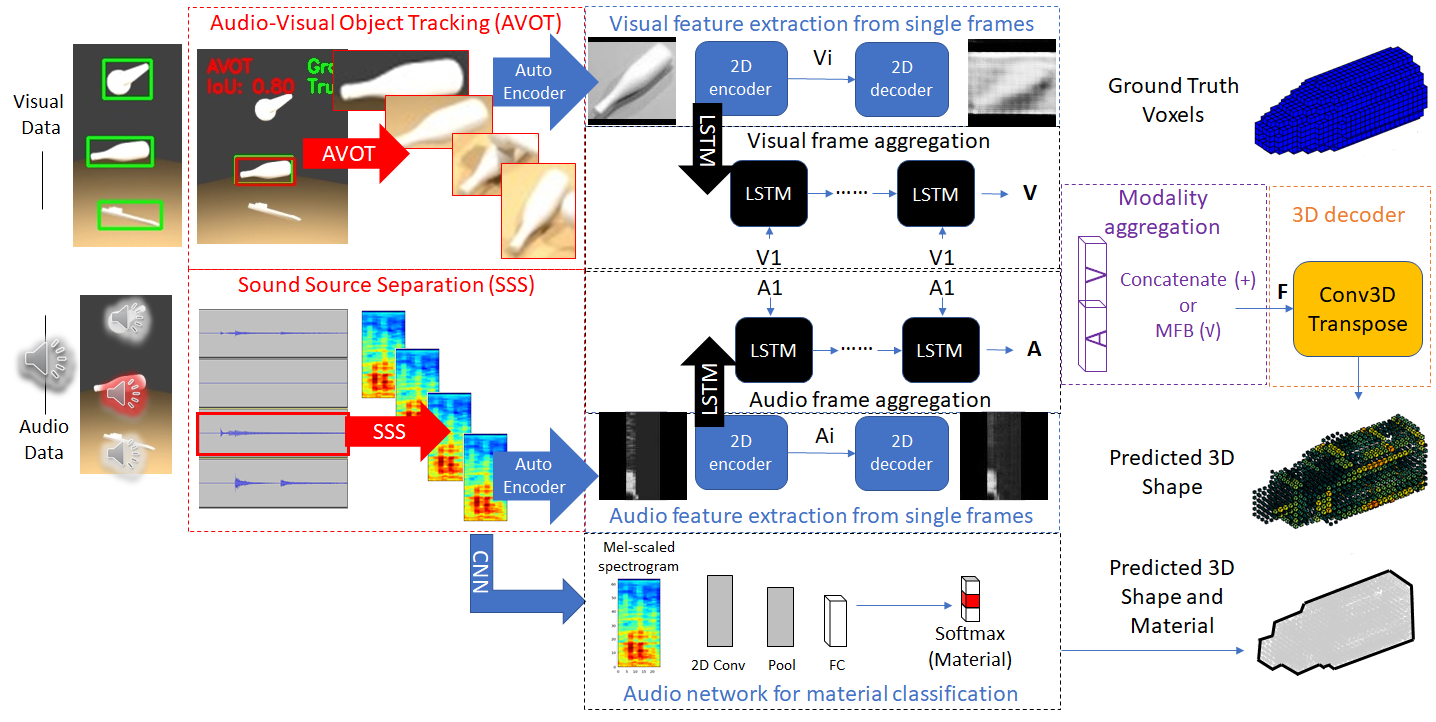}
  \caption{Our 3D-MOV neural network is a multimodal LSTM autoencoder optimized for 3D reconstructions of single ShapeNet objects and multiple objects from Sound20K video. During training, a LSTM autoencoder is trained to reconstruct 2D image and spectrogram inputs. 3D shape reconstructions are then generated by fine tuning the fused encodings of each modality for 3D voxel output. The network has recurrent LSTM layers for temporal consistency. Adding audio enhances learning for object tracking, material classification, and reconstruction when multiple objects collide, self-occlude, or are transparent.}
  \label{fig:fig1}
\end{figure*}

Main Results: In this paper, we introduce a new method to reconstruct high-quality 3D objects from video, as a sequence of images and sounds.  The main contributions of this work can be summarized as follows.
\begin{itemize}
\item A multimodal LSTM autoencoder neural network for both geometry and {\em material} reconstruction from audio and visual data is introduced;
\item The resulting implementation has been tested on voxel, audio, and image datasets of objects over a range of different geometries and materials;
\item Experimental results of our approach demonstrate the reconstruction of single sounding objects and multiple colliding objects in a virtual scene;
\item Audio-augmented datasets with ground-truth objects and their tracking bounding boxes are made available for research in audio-visual reconstruction from video.
\end{itemize}

\section{RELATED WORK}
\label{RelatedWork}

Computer vision research continues to push state-of-the-art reconstruction and segmentation of objects in a scene~\cite{DBLP:journals/corr/abs-1712-10215}. However, there still remain research opportunities in 3D reconstruction. Wide baselines limit the accuracy of feature correspondences between views. Challenging objects for reconstruction include thin or small objects (e.g. table legs), and classes of objects that are transparent, occluded, or have much higher shape variation than other classes (e.g. lamps, benches, and tables compared to cabinets, cars, and speakers for example). In this section, we review previous work relating to 3D reconstruction, multimodal neural networks, and reconstruction network structures.

\subsection{3D Reconstruction}

Deep learning techniques have produced state-of-the-art 3D scene and object reconstructions. These models take an image or series of images and generate a reconstructed output shape. Some methods produce a transformed image of the input, intrinsically representing the 3D object structure~\cite{Odena:17,Tsai:18,Mao:17,DBLP:journals/corr/MirzaO14,lun20173d}. 3D voxel grids provide a shape representation which is easy to visualize and works well with convolution operations~\cite{choy20163d,DBLP:journals/corr/GirdharFRG16,Riegler2017OctNet,qi2016volumetric,hu2018predictive,DBLP:journals/corr/0001ZXFT16}. In more recent work, point clouds have also been found to be a viable shape representation for reconstructed objects~\cite{Hedman:17,fan2017point}.

\subsection{Multimodal Neural Networks}
Neural networks with multiple modalities of inputs help cover a broader range of experimental setups and environments. Common examples include visual question answering~\cite{cadene2019murel}, vision and touch~\cite{lee2019making}, and other multisensory interactions~\cite{KLEMEN2012111}. Multiple modes may also take the form of image-to-image translation, e.g. domain transfer~\cite{Huang:18}. Using local and global cropped parts of the images (i.e.\ bounding boxes) have also been shown to serve as a mode of context to supervise learning~\cite{Reed:16}.

Audio-visual specific multimodal neural networks have also proven effective for speech separation~\cite{DBLP:journals/corr/abs-1804-03619} as well as sound localization~\cite{Zhao_2018_ECCV,DBLP:journals/corr/abs-1804-03641,Konno_2020_1533,ar2017look}. Audio synthesis conditioned on images is also enabled as a result of these combined audio-visual datasets~\cite{Zhou:18}. Please see a survey and taxonomy on multimodal machine learning~\cite{DBLP:journals/corr/BaltrusaitisAM17} and multimodal deep learning~\cite{10.5555/3104482.3104569} for more information.

\begin{figure*}
    \centering
  \includegraphics[width=0.9\textwidth]{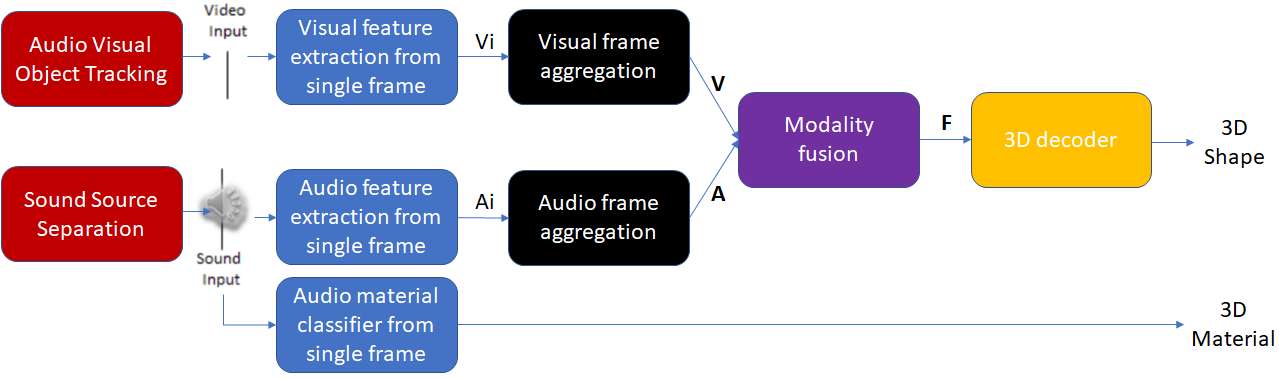}
  \caption{We first separate audio-visual data using object tracking (Section~\ref{VisualRepresentationAndSoundSourceSeparation}) and sound source separation (Section~\ref{AudioRepresentationAndSoundSourceSeparation}). Features from audio and visual subnetworks for each object are aggregated by LSTM autoencoders and then fused using addition, concatenation, or a bilinear model~\cite{yu2017mfb}. Finally, 3D geometry is reconstructed by a 3D decoder and audio classified material applied to all voxels.}
  \label{fig:technicalApproach}
\end{figure*}

\subsection{Reconstruction Network Structures}
While single view networks perform relatively well for most object classes, objects with concave structures or classes of objects with large variations in shape tend to require more views. 3D-R2N2~\cite{choy20163d} allows for both single and multi-view implementations given a single network. Other recurrent models include learning in video sequences~\cite{chong2017abnormal,hasan2016learning}, Point Set Generation~\cite{fan2017point}, and Pixel Recurrent Neural Network (PixelRNN)~\cite{DBLP:journals/corr/OordKK16}. Methods have also been developed to ensure temporal consistency~\cite{xie2018tempoGAN} and use generative techniques~\cite{DBLP:journals/corr/GwakCGCS17}. T-L network~\cite{DBLP:journals/corr/GirdharFRG16} and 3D-R2N2~\cite{choy20163d} are most similar to our 3D-MOV reconstruction neural network.  Building on these related works, we fuse audio as an additional input and %enforce 
temporal consistency in the form of LSTM layers (Fig.~\ref{fig:technicalApproach}).

\section{TECHNICAL APPROACH}
\label{TechnicalApproach}

In this work, we reconstruct the 3D shape and material of sounding objects given images and impact sounds. Using audio and visual information, we present a method for reconstruction of single instance ModelNet objects augmented with audio and multiple objects colliding in a Sound20K scene from video. In this section, we cover visual representations from object tracking (Section~\ref{VisualRepresentationAndSoundSourceSeparation}) and audio obtained from sound source separation of impact sounds (Section~\ref{AudioRepresentationAndSoundSourceSeparation}) that serve as inputs into our 3D-MOV reconstruction network (Section~\ref{NetworkStructure}).

\subsection{Object Tracking and Visual Representation}
\label{VisualRepresentationAndSoundSourceSeparation}
Since an entire video frame may contain too much background, we use object tracking to track and segment different objects. This tracking is performed using the Audio-Visual Object Tracker (AVOT)~\cite{Wilson2020}. Similar to the Single Shot MultiBox Detector (SSD)~\cite{DBLP:journals/corr/LiuAESR15}, AVOT is a feed-forward convolutional neural network that classifies and scales a fixed  number of anchor bounding boxes to track objects in a video. While 3D-MOV aggregates audio-visual features before decoding, AVOT fuses audio-visual inputs before its base network. With additional information from audio, AVOT defines an object based on both its geometry and material.

We use AVOT over other algorithms, such as YOLO~\cite{DBLP:journals/corr/RedmonDGF15} or Faster R-CNN~\cite{DBLP:journals/corr/RenHG015}, because of the availability of audio and need for higher object-tracking accuracy given occlusions caused by multiple objects colliding. Unlike CSR-DCF~\cite{DBLP:journals/corr/LukezicVCMK16}, AVOT automatically detects objects in the video without initial markup of bounding boxes. For future work, a scheduler network or a combination of object trackers is worth considering as well as use of Common Objects in Context (COCO)~\cite{DBLP:journals/corr/LinMBHPRDZ14} and SUN RGB-D~\cite{Song:15,Silberman:12,Janoch:11,Xiao:13} datasets for initialization and transfer learning. 

The output from tracking is a series of segmented image frames for each object, consisting of the contents of its tracked bounding box throughout the video. These segmented frames are grayscaled and resized to a consistent input size of 88 by 88 pixels. While resizing, we maintain aspect ratio and pad to square the image. These dimensions were automatically chosen to account for the size of objects in our Sound20K dataset and to capture their semantic information. Scenes included one, two, and three colliding objects with materials such as granite, slate, oak, and marble. For our single-frame, single impact sound evaluations, we resized ShapeNet's 224 x 224 image size. For comparison, other image sizes from related work include MNIST, 28 x 28; 3D-R2N2, 127 x 127; ImageNet, 256 x 256.

\begin{figure*}
  \centering
  \includegraphics[width=1\textwidth]{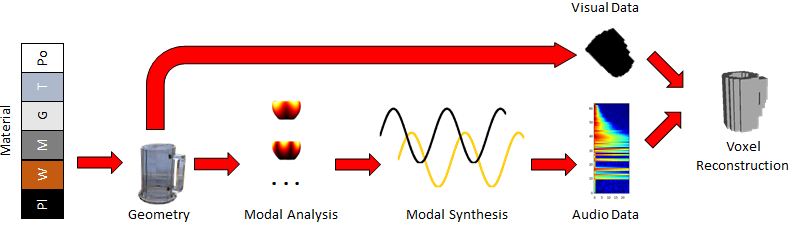}
  \caption{For our single impact sound analysis using ShapeNet, we build multimodal datasets using modal sound synthesis to produce spectrograms for audio input and images of voxelized objects as an estimate of shape. Please note that audio used from ISNN~\cite{sterling18isnn} was generated for voxelized models as a result of the sound synthesis pipeline requiring watertight meshes. Unmixed Sound20K audio was available from the generated synthetic videos.}
  \label{fig:singleViewData}
\end{figure*}

\begin{figure*}
  \centering
  \includegraphics[width=0.9\textwidth]{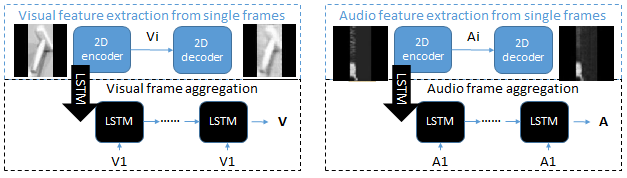}
  \caption{Hidden layer representations $V_i$ and $A_i$ are trained to spatially encode object geometry and impact sounds, where i is each video frame. These learned weights are subsequently used during test time to generate 3D shapes from audio-visual inputs. For sequence modeling, LSTM layers are reliable for temporal consistency and establishing dependencies over time. More specifically, we use convolutional LSTM layers rather than fully connected to also preserve spatial information.
  }
  \label{fig:lstmAE}
\end{figure*}

\subsection{Sound Source Separation of Impact Sounds and Audio Representation}
\label{AudioRepresentationAndSoundSourceSeparation}
For single frame reconstruction, we synthesize impact sounds on ShapeNet~\cite{sterling18isnn}, illustrated in Fig.~\ref{fig:singleViewData}. For multiple frames, we take as input a Sound20K video showing one or more objects moving around a scene. These objects strike one another or the environment, producing impact sounds, which can be heard in the audio track of the video. We refer to these objects, dynamically moving through the scene and generating sound due to impact and collision, as \emph{sounding objects}. Sound20K provides mixed and unmixed audio which can be used directly or to train algorithms for sound source separation~\cite{Wang:14,Koretzky:17,Scallie:17}. While prior work to localize objects using audio-visual data exists~\cite{ar2017look,Zhao_2018_ECCV}, automatically associating separated sounds with corresponding visual object tracks in the context of the reconstruction task remains an area of future work.

Initially, Sound20K and ShapeNet audio are available as time series data, sampled at 44.1 kHz to cover the full audible range. The audio is converted to mel-scaled spectrograms for neural network inputs, which effectively represent the spectral distribution of energy over time.
Each spectrogram is 3 seconds for a single frame (ShapeNet) and 0.03 seconds per multi-frame (Sound20K) with an overlap of 25\%. Audio spectrograms are aligned temporally with their corresponding image frames from video, forming the audio-visual input for queries. They are generated with %FFTs
discrete short-time Fourier transforms (STFTs) using a Hann window function.

\begin{equation}
    \chi(m,k) = \sum^{N-1}_{n=0} x(n+mH)w(n)exp(-2\pi ikn/N)
\end{equation}
\label{eqn:stft}

\noindent for $m^{th}$ time frame and $k^{th}$ Fourier coefficient with real-valued DT signal $x: Z \rightarrow R$, sampled window function $w(n)$ for n $\in [0:N-1] \rightarrow R$ of length $N \in N$, and hop size $H \in N$~\cite{M15}.

\subsubsection{Single View, Single Impact Sound}
Single-view inputs are based on ShapeNet, a repository of 3D CAD models based on WordNet categories. Evaluations were performed on voxelized versions of ShapeNet's~\cite{changShapeNet15}, ModelNet10 and ModelNet40 models~\cite{shapenetCVPR15}, and image views of these datasets from 3D-R2N2~\cite{choy20163d}. To generate audio for these objects to be used for our multi-modal 3D-MOV neural network, we use data from Impact Sound Neural Network~\cite{sterling18isnn}. This work synthesized impact sounds for voxelized ModelNet10 and ModelNet40 models~\cite{shapenetCVPR15} using modal analysis and sound synthesis. Modal analysis is precomputed to obtain \textit{modes} of vibration for each object and sound synthesized with an amplitude determined at run-time given the hit point location on the object and impulse force. The modes are represented as damped sinusoidal waves where each mode has the form
\begin{equation}
	q\textsubscript{i} = a\textsubscript{i}e\textsuperscript{-d\textsubscript{i}t}sin(2\pi f\textsubscript{i}t + \theta\textsubscript{i}),
\end{equation}
\noindent
where $f\textsubscript{i}$ is the frequency of the mode, $d\textsubscript{i}$ is the damping coefficient, $a\textsubscript{i}$ is the excited amplitude, and $\theta \textsubscript{i}$ is the initial phase.

\subsubsection{Multi-Frame, Multi-Impact}
Multi-frame inputs to our system consist of Sound20K~\cite{gensound} videos that may contain multiple sounding objects, possibly of similar sizes, shapes, and/or materials. This synthetic video dataset contains audio and video data for multiple objects colliding in a scene. Sound20K consists of 20,378 videos generated by rigid-body simulation and impact sound synthesis pipeline~\cite{femJames06}. Visually, Sound20K~\cite{gensound} objects can be separated from one another through tracking of bounding boxes. However, audio source separation can be more challenging, particularly for unknown objects. While Sound20K provides separate audio files for each object that can be used, the audio data can also be used to train sound source separation techniques~\cite{Wang:14,Koretzky:17,Scallie:17} to learn to unmix audio to individual objects by geometry and material. 
As future work, we will compare the impact on reconstruction quality and performance if we were to use combined, unmixed audio for each object. We will also compare impact of using source separated sounds versus ground truth unmixed audio.

\begin{figure*}
  \centering
  \includegraphics[width=1\textwidth]{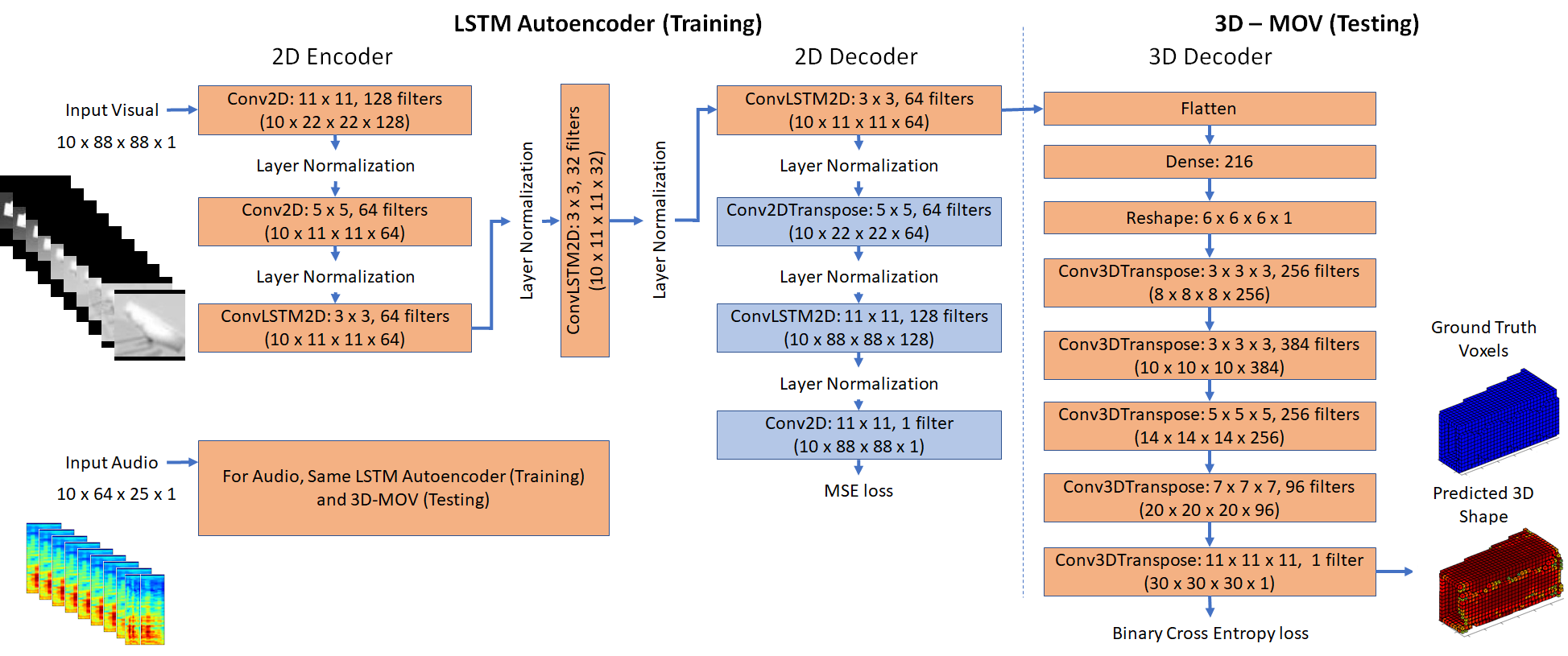}
  \caption{We separately train audio and visual autoencoders to learn encodings and fine-tune for our 3D reconstruction task. We replace the 2D decoder by a five deconvolutional layer 3D decoder to generate a $30^3$ voxel grid. The separate audio-visual LSTM autoencoders are flattened and merged to form the dense layer. Here, the predicted 3D shape voxels are displayed based on a threshold of 0.3.}
  \label{fig:networkStructure}
\end{figure*}

\section{3D-MOV NETWORK STRUCTURE}
\label{NetworkStructure}

Our 3D-MOV network is a multi-modal LSTM autoencoder optimized for 3D reconstructions of multiple objects from video. Like 3D-R2N2~\cite{choy20163d}, it is recurrent and generates a 3D voxel representation. However, to the best of our knowledge, our 3D-MOV network is the first audio-visual reconstruction network for 3D object reconstruction. After object tracking and sound source separation, we separately train autoencoders to extract visual and audio feature from each frame (Section~\ref{sec:SingleFrameAudioVisualFeatureExtraction}). While the 2D encoder weights are reused, the 2D decoders are discarded (blue rectangles in Fig.~\ref{fig:networkStructure}) and replaced with 3D decoders for learning to reconstruct voxel outputs of the tracked objects based on given 2D images and spectrograms.
Using a merge layer such as addition, concatenation, or a bilinear model~\cite{yu2017mfb}, our method 3D-MOV fuses the results of the audio and visual subnetworks comprised of LSTM autoencoders.

\subsection{Single Frame Feature Extraction}
\label{sec:SingleFrameAudioVisualFeatureExtraction}
The autoencoder consists of two convolutional layers for spatial encoding followed by a LSTM convolutional layer for temporal encoding. As a general rule of thumb, we use small filters (3x3 and at most 5x5), except for the very first convolutional layer connected to the input, and strides of four and two for the two conv layers~\cite{cs231n}. %Because the problem is symmetric, the mechanism that encodes the first object images should be reused (weights and all) to encode the second object images. 
The decoder mirrors the encoder to reconstruct the image (Fig.~\ref{fig:lstmAE}). After each convolutional layer, we employ layer normalization, which is equivalent to batch normalization for recurrent networks~\cite{ba2016layer}. It normalizes the inputs across features and is defined as:

\begin{equation}
    \mu_{j} = \frac{1}{m}\sum_{j=1}^m x_{ij}; \sigma_j^2 = \frac{1}{m} \sum_{j=1}^m (x_{ij} - \mu_j)^2; \hat{x}_{ij} = \frac{x_{ij}-\mu_j}{\sqrt{\sigma_j^2 + \epsilon}}
\end{equation}

\noindent where $x_{ij}$ is batch i, feature j of the input x across m features.

\subsection{Frame Aggregation}
\label{FrameAggregation}
In chronological order, the training video frames make a temporal sequence. LSTM convolutional layers are used to preserve content and spatial information. To generate more training sequences, we perform data augmentation by concatenating frames with strides 1, 2, and 3. For example, we use a skipping stride of 2 to generate a sequence of every other frame. We use a 10-frame sequence size as a sliding window technique for aggregation of the encodings. The encoder weights learned here are used to then learn 3D decoder weights to output a 3D voxel reconstruction based on audio-visual inputs from audio-augmented ModelNet with impact sound synthesis and Sound20K video.

\subsection{Modality Fusion and 3D Decoder}
After encoding our inputs with LSTM convolutional layers, we flatten to a fully connected layer for each audio and visual subnetwork. These dense layers are fused together prior to multiple Conv3D transpose layers for the 3D decoder. Prior work in multimodal deep learning, such as visual question and answering systems, have merged modalities for classification tasks using addition and MFB~\cite{yu2017mfb}. %These techniques have also been demonstrated in visual question and answering systems. 
A 3D decoder accepts the fusion of audio-visual LSTM encodings and maps it to a voxel grid with five deconvolutional layers, similar to T-L Network~\cite{DBLP:journals/corr/GirdharFRG16}. %However, u
Unlike T-L's $20^{3}$ voxel grid, we use $30^{3}$ voxels for greater resolution and apply a single, audio-based material classification to all voxels. Deconvolution, also known as fractionally-strided or transposed convolution, results in a %larger output
3D voxel shape by broadcasting input $X$ through kernel $K$~\cite{zhang2020dive}.

\begin{equation}
    \sum_{i=0}^h \sum_{j=0}^w Y[i: i + h,j: j + w] \mathrel{+}= (X[i, j] * K)
\end{equation}

\section{RESULTS}

\label{Results}

\begin{figure*}%[ht]
  \centering
  \includegraphics[width=0.8\textwidth]{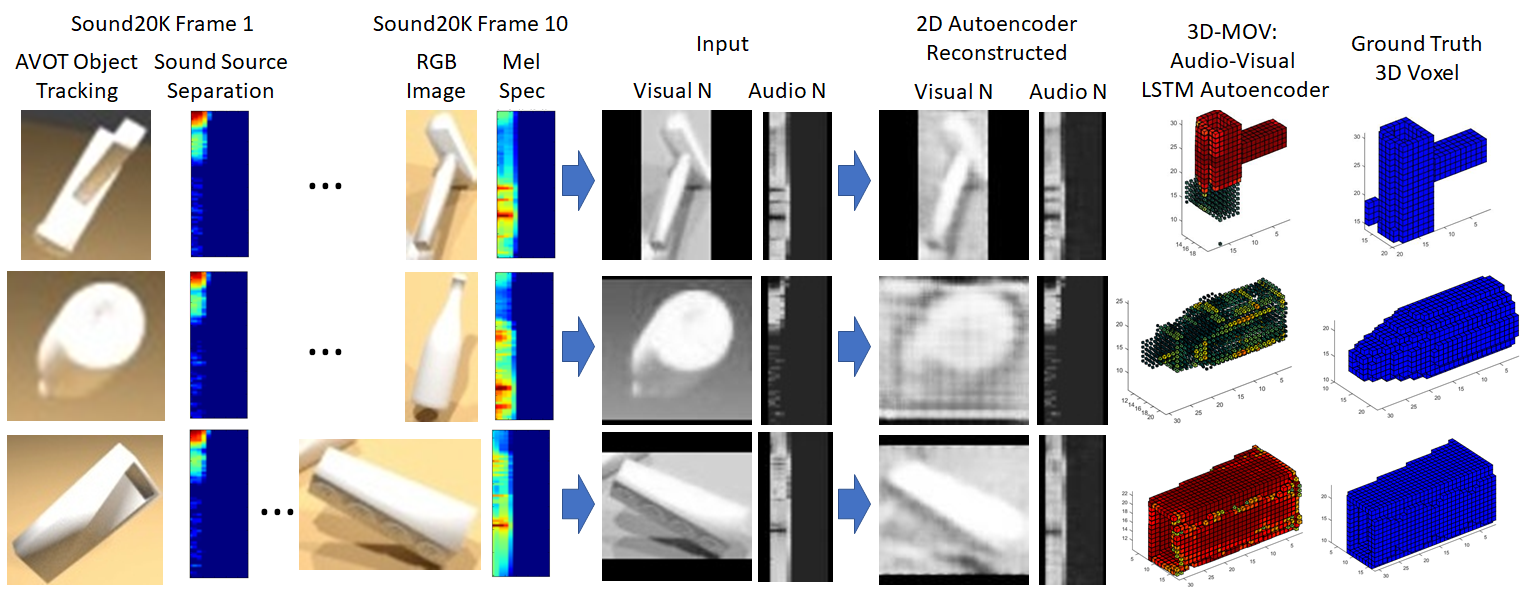}
  \caption{Reconstructed objects from using multiple frames and impact sounds. Please see our supplementary materials for a complete review of results for ShapeNet and Sound20K datasets using binary cross entropy loss and reconstruction accuracy comparing audio, visual, and audio-visual methods by number of views. Our method is able to obtain better reconstruction results for concave internal structures and scenes with multiple objects by fusing temporal audio-visual inputs.
  }
  \label{fig:results}
\end{figure*}

In this section, we present our implementation, training, and evaluation metrics along with 3D-MOV reconstructed objects (Fig.~\ref{fig:results}). Please see the accompanying supplementary materials for more comparative analysis of loss and accuracy against baseline methods by datasets and numbers of views. For each of ShapeNet and Sound20K, we evaluate the network architecture in Section~\ref{NetworkStructure} against audio, visual, and audio-visual methods using binary cross entropy loss and intersection over union (IoU) reconstruction accuracy.

\subsection{Implementation}
\label{Implementation}

Our framework was implemented using Tensorflow~\cite{tensorflow2015-whitepaper} and Keras~\cite{chollet2015keras}. Training was run on Ubuntu 16.04.6 LTS with a single Titan X GPU. Voxel representations were rendered based on Matlab visualization code from 3D-GAN~\cite{DBLP:journals/corr/0001ZXFT16}. From Sound20K videos, images were grayscale with dimensions 84 x 84 x 1 and audio spectrograms were 64 x 25 x 1, zero padded to equivalent dimensions. Visual data was augmented with resizing, cropping, and skipping strides.

\begin{figure}
  \centering
  \includegraphics[width=0.45\textwidth]{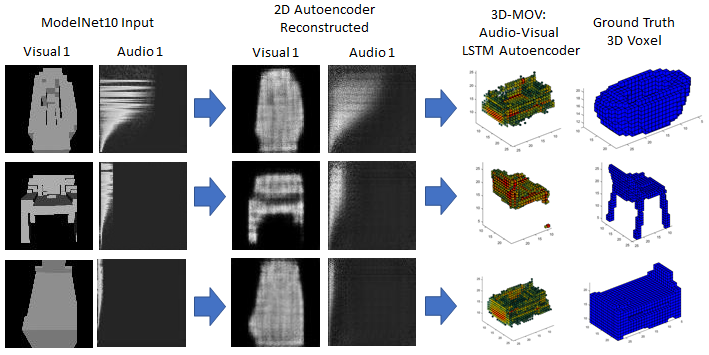}
  \caption{3D-MOV-AV reconstructed image and audio inputs for single view voxelized ModelNet10 classes (top, bathtub; middle, chair; bottom, bed). These results are %using a single image and single impact sound, 
  fusing a single image and impact sound with an addition merge layer, training for 60 epochs on a single GPU, and using a voxel threshold of 0.4. 3D-MOV-AV performs best on ModelNet10 single views with audio augmenting visual data.}
  \label{fig:3dmov_av}
  % \vspace*{-1em}
\end{figure}

\subsection{Training}
\label{Training}

Since joint optimization can be difficult to perform, we train our reconstruction autoencoder and fused audio-visual networks separately and then jointly optimize to fine-tune the final network. Mean square error is used for the 2D reconstruction loss to train the encoder to reconstruct input images and audio spectrograms. Binary cross entropy loss is calculated between ground truth and reconstructed 3D voxel grids. During testing, we reconstruct from encoded vector representation of audio-visual inputs to a 3D voxel reconstruction output.

Previous work has used symmetry induced volume refinement to constrain and finalize GAN volumetric outputs~\cite{DBLP:journals/corr/abs-1804-05469}. Other methods have used multiple views to continuously refine the output~\cite{choy20163d}. Furthermore, most adversarial generating methods create examples by perturbing existing data, limiting the solution space. Our approach constrains the space of possible 3D reconstructions for objects in the scene by temporal consistency, aggregation, and fusion of audio and visual inputs.

\setlength{\tabcolsep}{4pt}
\begin{table}
\begin{center}
\begin{tabular}{lc|cc}
\hline\noalign{\smallskip}
Dataset & & \multicolumn{2}{c}{ShapeNet~\cite{changShapeNet15}} \\
Method & Input & 1 view & 5 views \\
\noalign{\smallskip}
\hline
\noalign{\smallskip}
\textbf{3D-MOV-A (Ours)} & A & 21.2\% & N/A \\
\hline
% 3D-R2N2~\cite{choy20163d} & V & 56.0\%** & 63.1\%** \\
\textbf{3D-MOV-V (Ours)} & V & 22.7\% & 22.5\% \\
\hline
T-L Network~\cite{DBLP:journals/corr/GirdharFRG16} & AV & 18.0\%* & N/A\\
\textbf{3D-MOV-AV (Ours)} & AV & {\bf 32.6\%} & {\bf 31.0\%} \\
\hline
\end{tabular}
\end{center}
\end{table}
\setlength{\tabcolsep}{1.4pt}

\setlength{\tabcolsep}{4pt}
\begin{table}
\begin{center}
\begin{tabular}{lc|c}
\hline\noalign{\smallskip}
Dataset & & \multicolumn{1}{c}{Sound20K~\cite{gensound}} \\
Method & Input & 10 views \\
\noalign{\smallskip}
\hline
\noalign{\smallskip}
\textbf{3D-MOV-A (Ours)} & A & 37.2\% \\
\hline
% 3D-R2N2~\shortcite{choy20163d} & V & TBU & N/A \\
\textbf{3D-MOV-V (Ours)} & V & 65.7\% \\
\hline
%T-L Network~\shortcite{DBLP:journals/corr/GirdharFRG16} & AV & TBU & N/A\\
%\textbf{3D-MOV-AV (Ours)} & AV & 64.6\% & 69.8\% \\
\textbf{3D-MOV-AV (Ours)} & AV & {\bf 69.8\%} \\
\hline
\end{tabular}
\end{center}
\caption{3D-MOV was evaluated for loss %against baselines for loss %(mean square error and binary cross entropy) 
and reconstruction accuracy. %(intersection over union).
A view consists of an image and audio frame. Very slight decreases in 3D-MOV accuracy for 1 or 5 ShapeNet views 
%increase requires further investigation but
may suggest impact sounds of different hit points are needed rather than using the same sound across views. *We use the T-L Network~\cite{DBLP:journals/corr/GirdharFRG16} fused with audio as an overall baseline comparison.
3D-MOV-AV shows performance improvement over single-modality input on both datasets.  % with 0.67 loss and 18.0\% IoU for an instance of the MN10 chair class. 
% ** Reported in~\cite{choy20163d}}
}
\vspace*{-3em}
\label{table:results}
\end{table}
\setlength{\tabcolsep}{1.4pt}

\subsection{Evaluation metrics}
Methods were evaluated against voxel Intersection-over-Union (IoU), also known as the Jaccard index~\cite{Jaccard1901}, between the 3D reconstruction and ground truth voxels as well as cross-entropy loss. This can be represented as area of overlap divided by the area of union. More formally:

\begin{equation}
    IoU = \frac{\sum_{i,j,k} [I(p_{(i,j,k)} > t)I(y_{(i,j,k)}]}{\sum_{i,j,k} [I(I(p_{(i,j,k)} > t) + I(y_{(i,j,k)}))]}
\end{equation}
\label{eqn:IoU}

\noindent where $y_{i,j,k} \in {0,1}$ is the ground truth occupancy, $p_{i,j,k}$ the Bernoulli distribution output at each voxel, $I(\cdot)$ an indicator function, and t for threshold. Higher IoU means better reconstruction quality.

\section{CONCLUSIONS}
\label{Conclusions}

To the best of our knowledge, this work is the first method to use audio and visual inputs from ShapeNet objects and Sound20K video of multiple objects in a scene to generate 3D object reconstructions {\em with material} from video. While multi-view approaches can improve reconstruction accuracy, transparent objects, interior concave structures, self-occlusions, and multiple objects remain a challenge. As objects collide, audio provides a complementary sensory cue that can enhance the reconstruction model to improve results. In this paper, we demonstrate that augmenting image encodings with corresponding impact sounds refine reconstructions of multimodal LSTM autoencoder neural network outputs.

\noindent
{\bf Limitations:} our approach is currently implemented and evaluated with fixed-grid shapes. Further experimentation with residual architectures~\cite{DBLP:journals/corr/HeZRS15}, adaptive grids, and multi-scale reasoning~\cite{DBLP:journals/corr/DentonCSF15} are worth exploring, though they each will introduce different sets of constraints and complexity. Material classification is predicted based on audio alone, given the textureless image renderings of the datasets used. Also, only a single material is inferred for the entire geometry rather than per voxel classification. Finally, the trade-off between additional views and additional auditory inputs could be further explored.

\noindent
{\bf Future Work:} evaluation of other real-time object trackers, such as YOLO and Faster R-CNN, can be performed and trained on other existing datasets, such as COCO and SUN RGB-D. Further investigations can also examine how the error introduced by object tracking propagates to reconstruction error. Same applies to errors from sound source separation and being able to accurately associate unmixed sounds with their corresponding visual object tracks. Next, while audio helps classify the material of the reconstructed geometry, we assume a single material classification based on audio alone and apply that to all voxels. Research on classifying material per voxel using both audio and visual data could expand part segmentation research into reconstructing objects with different materials. Rather than being fully deterministic, fusing audio and visual information for generative models to reconstruct geometry and material may also be of interest to the research community. Then, there may be more than one possible 3D reconstruction for a given image or sound. Beyond reconstruction, audio may also enhance image and sound generation, as well as memory and attention models. For instance, image generation using an audio conditioned GAN and sound generation based on image conditioning could be explored, similar to WaveNet~\cite{Oord:16} local and global conditioning techniques. Finally, testing on real data in the wild and larger datasets of annotated audio and visual data allow for future research. %continued research in this area.


%%%%%%%%%%%%%%%%%%%%%%%%%%%%%%%%%%%%%%%%%%%%%%%%%%%%%%%%%%%%%%%%%%%%%%%%%%%%%%%%

%%%%%%%%%%%%%%%%%%%%%%%%%%%%%%%%%%%%%%%%%%%%%%%%%%%%%%%%%%%%%%%%%%%%%%%%%%%%%%%%

%%%%%%%%%%%%%%%%%%%%%%%%%%%%%%%%%%%%%%%%%%%%%%%%%%%%%%%%%%%%%%%%%%%%%%%%%%%%%%%%
%\section*{APPENDIX}


%\section*{ACKNOWLEDGMENT}

%The preferred spelling of the word ÒacknowledgmentÓ in America is without an ÒeÓ after the ÒgÓ. Avoid the stilted expression, ÒOne of us (R. B. G.) thanks . . .Ó  Instead, try ÒR. B. G. thanksÓ. Put sponsor acknowledgments in the unnumbered footnote on the first page.

%%%%%%%%%%%%%%%%%%%%%%%%%%%%%%%%%%%%%%%%%%%%%%%%%%%%%%%%%%%%%%%%%%%%%%%%%%%%%%%%

%References are important to the reader; therefore, each citation must be complete and correct. If at all possible, references should be commonly available publications.

\bibliographystyle{IEEEtran}
\bibliography{IEEEabrv,IEEEexample}

\end{document}